\documentclass{article}
\newenvironment{ack}{\section*{Acknowledgments}}{}

\usepackage{arxiv}

\usepackage[utf8]{inputenc} 
\usepackage[T1]{fontenc}    
\usepackage{hyperref}       
\usepackage{url}            
\usepackage{booktabs}       
\usepackage{amsfonts}       
\usepackage{amsthm}         
\usepackage{nicefrac}       
\usepackage{microtype}      
\usepackage{natbib} 
\usepackage{lipsum}
\usepackage{graphicx}
\graphicspath{ {./images/} }

\usepackage{amsmath,amsfonts,amssymb}
\usepackage{tabularx}
\usepackage{threeparttable}
\usepackage{comment}
\usepackage{algorithm}
\usepackage{algorithmic}

\newtheorem{theorem}{Theorem}
\newtheorem{lemma}[theorem]{Lemma}

\title{A Contrastive Diffusion-based Network (CDNet) for Time Series Classification}

\author{
 Yaoyu Zhang \\
  Department of Mechanical and Industrial Engineering\\
  University of Toronto\\
  \texttt{yaoyu.zhang@mail.utoronto.ca} \\
   \And
 Chi-Guhn Lee \\
  Department of Mechanical and Industrial Engineering\\
  University of Toronto\\
  \texttt{chiguhn.lee@utoronto.ca} \\
}

\begin{document}

\maketitle

\begin{abstract}
Deep learning models are widely used for time series classification (TSC) due to their scalability and efficiency. However, their performance degrades under challenging data conditions such as class similarity, multimodal distributions, and noise. To address these limitations, we propose CDNet, a \textit{Contrastive Diffusion-based Network} that enhances existing classifiers by generating informative positive and negative samples via a learned diffusion process. Unlike traditional diffusion models that denoise individual samples, CDNet learns transitions between samples—both within and across classes—through convolutional approximations of reverse diffusion steps. We introduce a theoretically grounded CNN-based mechanism to enable both denoising and mode coverage, and incorporate an uncertainty-weighted composite loss for robust training. Extensive experiments on the UCR Archive and simulated datasets demonstrate that CDNet significantly improves state-of-the-art (SOTA) deep learning classifiers, particularly under noisy, similar, and multimodal conditions.
\end{abstract}


\section{Introduction}

\begin{figure}[ht]
\centering
\includegraphics[width=0.5\textwidth]{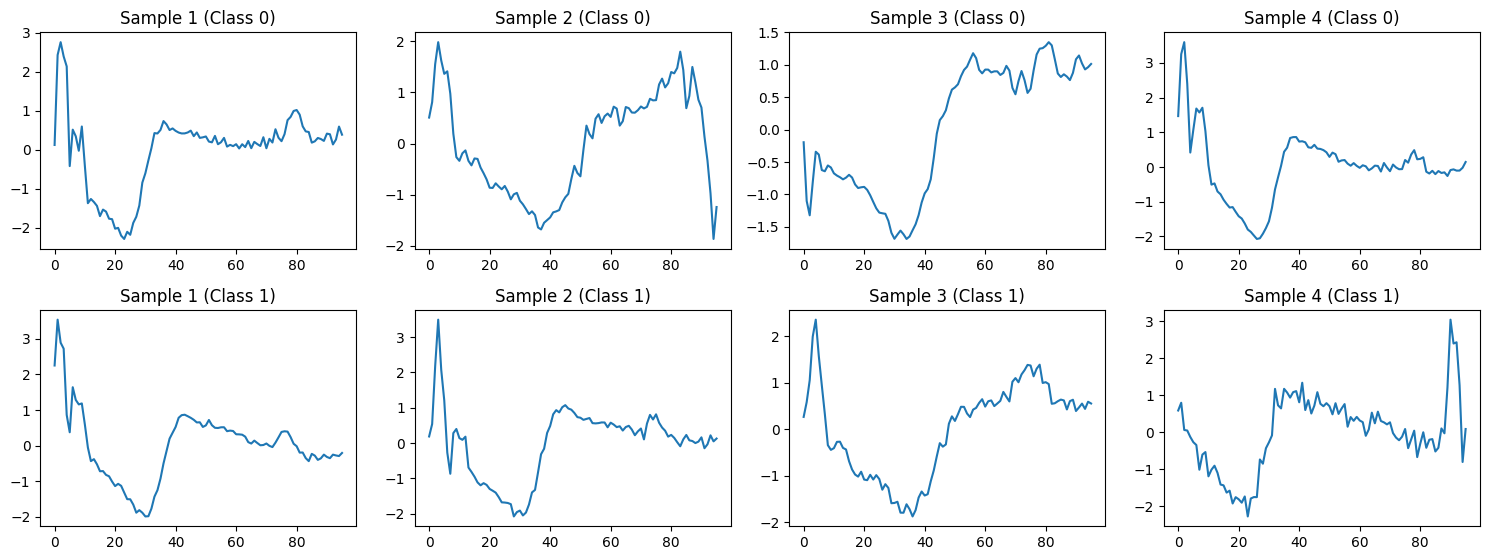}
\caption{The ECG200 dataset from the UCR Archive. The top row and bottom row correspond to two classes, respectively. Two classes generally have similar shapes, but various patterns can be found within each class.}
\label{fig:sample_ecg}
\end{figure}

Time series classification (TSC) is foundational to many real-world applications in healthcare, industrial systems, and finance \cite{lim2021time, masini2023machine}. Deep learning models, particularly convolutional and recurrent neural networks, have become the dominant choice for TSC due to their scalability and capacity for end-to-end learning \cite{ismail2019deep}. However, their performance is often unreliable in realistic scenarios where time series exhibit noise, high inter-class similarity, and multimodal intra-class variability. These challenges frequently arise in domains like ECG monitoring, where samples from different classes may look deceptively similar, while samples from the same class vary significantly in shape (Figure~\ref{fig:sample_ecg}) \cite{dau2019ucr}.

In such complex settings, state-of-the-art (SOTA) deep TSC models—despite their scalability—often underperform compared to classical methods \cite{dhariyal2020examination}. This performance gap is especially notable in univariate settings where expressive representations are limited. Moreover, most SOTA non-deep methods, including shape-based, feature-based, and dictionary-based techniques, are not well-suited for tasks like transfer learning or deployment in large-scale systems \cite{bolya2021scalable}. Thus, a key research challenge lies in enhancing deep classifiers so they can better handle noisy, similar, and multimodal time series distributions without sacrificing scalability.

Existing TSC approaches span several families: dictionary-based models like WEASEL \cite{schafer2017fast} and BOSS \cite{middlehurst2019scalable} discretize sequences into histograms; distance-based methods such as DTW \cite{berndt1994using} and ShapeDTW \cite{zhao2018shapedtw} rely on alignment; feature-based models like TSFresh \cite{christ2018time} and FreshPRINCE \cite{middlehurst2022freshprince} extract interpretable summary statistics; and deep learning models like CNNs \cite{sadouk2019cnn}, LSTMs \cite{karim2019multivariate}, and MultiRocket \cite{tan2022multirocket} are trained end-to-end. While ensemble methods further boost robustness, they typically demand more data and computation. Ultimately, few of these methods address the core challenge of ambiguous or highly variable sample distributions.

To resolve these issues, we draw inspiration from diffusion models—originally developed for generative modeling in vision \cite{ho2020denoising}—and adapt them to TSC. Diffusion models learn to reverse a process of gradually adding Gaussian noise, and have been extended for tasks beyond generation, such as classification with priors \cite{han2022card}. Parallel to this, contrastive learning frameworks like BYOL \cite{grill2020bootstrap}, Barlow Twins \cite{zbontar2021barlow}, and recent time series–specific methods such as TS-TCC \cite{demirel2023tstcc}, TC-TCC \cite{eldele2021tctcc}, TS2Vec \cite{yue2022ts2vec}, and T-Loss \cite{wang2021tloss} have shown strong performance by leveraging augmentations to build contrastive pairs for representation learning. However, these methods typically rely on heuristic or random augmentations and do not explicitly model structured transitions between samples.

In this work, we propose the Contrastive Diffusion-based Network (CDNet), a framework that augments a base classifier with a learned diffusion process that generates meaningful transitions between samples. Unlike traditional diffusion models that denoise an input to its original state, CDNet performs transitions \emph{between} instances—both within and across classes—using a CNN-based approximation of the reverse process. This enables the model to learn intra-class variability, inter-class separability, and denoising simultaneously, all while generating sample trajectories that are semantically grounded and useful for contrastive training.

\begin{figure*}[ht]
\centering
\includegraphics[width=\textwidth]{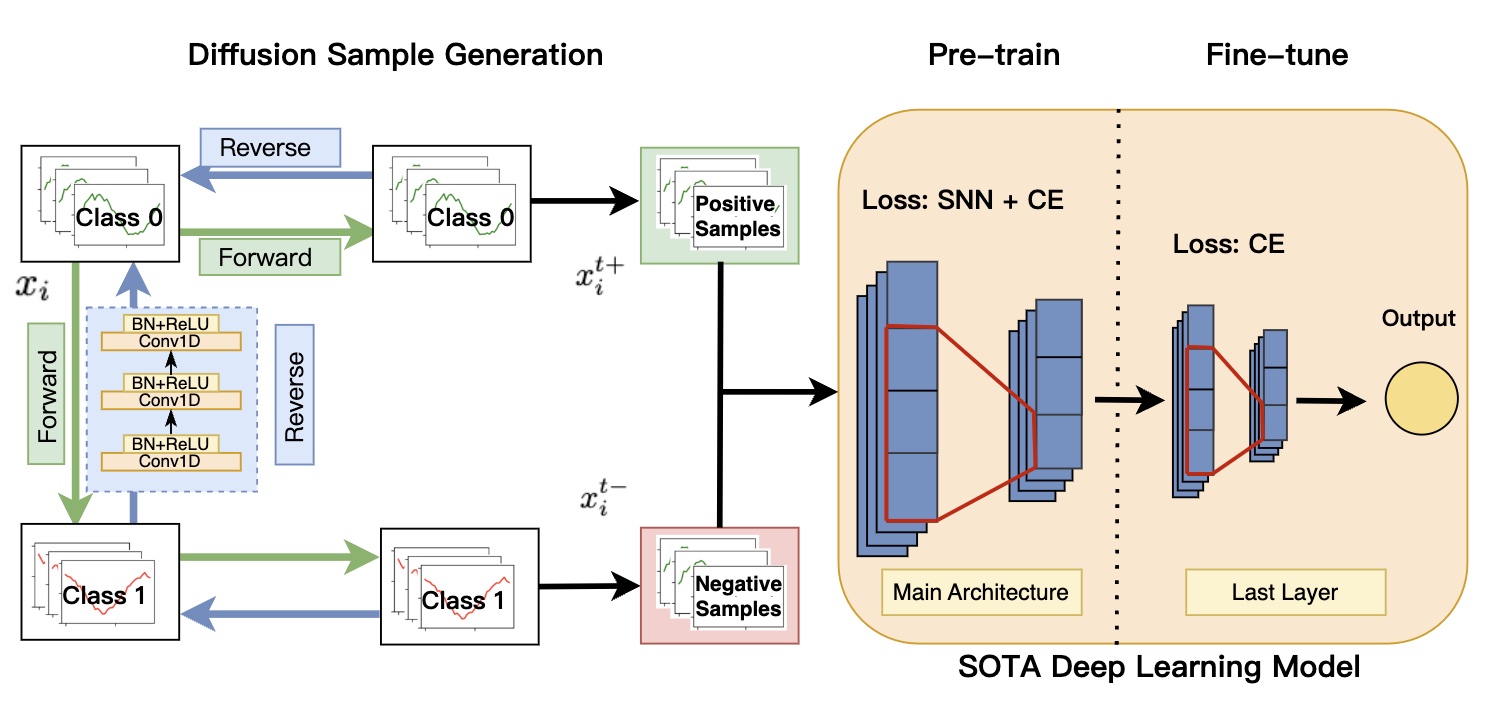}
\caption{An overview of CDNet.}
\label{fig:cdnet}
\end{figure*}

As shown in Figure~\ref{fig:cdnet}, CDNet is built in two stages. In the pre-training stage, we generate positive and negative samples for each input by learning transitions via a reverse diffusion process implemented with CNNs. This allows us to build contrastive pairs that capture variability within classes and similarity across classes under noisy conditions. We pre-train with a composite of cross-entropy loss, triplet loss, and soft nearest neighbor loss. To effectively balance these objectives, we introduce an uncertainty-weighted loss that adjusts dynamically during training. In the fine-tuning stage, we freeze earlier layers and retrain only the classifier head using the original labels.

We validate CDNet on the UCR Time Series Archive \cite{dau2019ucr}, focusing on binary univariate datasets where ambiguity is most problematic. We show that applying CDNet on top of leading deep classifiers—InceptionTime (IT) \cite{ismail2020inceptiontime}, 1DCNN \cite{zhao2017convolutional}, and LSTM\_FCN \cite{karim2019multivariate}—significantly improves accuracy. We also conduct a simulation study to examine how CDNet's gains vary with noise, inter-class similarity, and multimodality.

\vspace{0.2em}
\noindent \textbf{Our key contributions are:}
\begin{itemize}
    \item We propose CDNet, a novel diffusion-based augmentation framework that learns transitions between time series instances—within and across classes—to generate contrastive sample trajectories for robust representation learning.
    
    \item We establish theoretical guarantees showing that 1D CNNs can approximate reverse diffusion transitions with arbitrarily low error, enabling effective denoising and support recovery for multimodal distributions.

    \item We introduce an uncertainty-weighted composite loss that adaptively balances classification and contrastive objectives during pre-training, improving robustness under noisy and ambiguous conditions.

    \item We empirically demonstrate that CDNet consistently improves the performance of state-of-the-art deep TSC models (e.g., InceptionTime, 1DCNN, LSTM\_FCN) on the UCR Archive and simulated datasets, particularly under high noise, inter-class similarity, and intra-class variability.
\end{itemize}

\section{Related Work}
In addition to the mainstream SOTA time series classification models mentioned in the Introduction section, the proposed model is developed by innovating from the naive diffusion process defined as follows \cite{croitoru2023diffusion}. Given a univariate time series $x$, a diffusion model includes a forward process that gradually adds noise to the time series over several state steps $t$ and is defined as follows:
$$
x^{t}=\sqrt{1-\beta^t} x^{t-1}+\sqrt{\beta^t} \epsilon
$$
where $\beta^t$ is control the rate of noise added at each state step $t$, and $\epsilon$ is Gaussian noise.

The goal of the forward process is to inject noise starting from $x^0$ at state step $t \in \{1,..,T\}$. The distribution of $x^t$ given the previous state step $x^{t-1}$ is

$$
q\left(x^t \mid x^{t-1}\right)=\mathcal{N}\left(x^t ; \sqrt{1-\beta^t} x^{t-1}, \beta^t I\right)
$$

On the contrary, the reverse process focuses on removing the Gaussian noise injected from the forward process, starting from $x^T$ back to $x^0$ :

$$
x^{t-1}=\mu_\theta\left(x^t, t\right)+\sigma_\theta \cdot \epsilon
$$
where $\mu_\theta\left(x^t, t\right)$ and $\sigma_\theta$ are predicted by the model, and $\epsilon \sim \mathcal{N}(0, I)$. The distribution of the reverse process parameterized by $\theta$ is given by:

$$
p_\theta\left(x^{t-1} \mid x^t\right)=\mathcal{N}\left(x^{t-1} ; \mu_\theta\left(x^t, t\right), \sigma_\theta^2 I\right)
$$

The training objective is to minimize the variational bound on the data likelihood.

The naive diffusion model can be further extended for classification and regression by adapting knowledge learned from a pre-trained model $f_{\phi}(x)$, with the reverse process defined as follows \cite{han2022card}:

$$
p\left(y^T \mid x\right)=\mathcal{N}\left(f_\phi(x), I\right),
$$

and the forward process is defined as follows:

\begin{align*}
& q\left(y^t \mid y^{t-1}, f_\phi(x)\right)\\
& =\mathcal{N}\left(y^t ; \sqrt{1-\beta^t} y^{t-1}+\left(1-\sqrt{1-\beta^t}\right) f_\phi(x), \beta^t I\right).
\end{align*}

\section{Method}\label{s: method}

\begin{figure*}[ht]
\centering
\includegraphics[width=\textwidth]{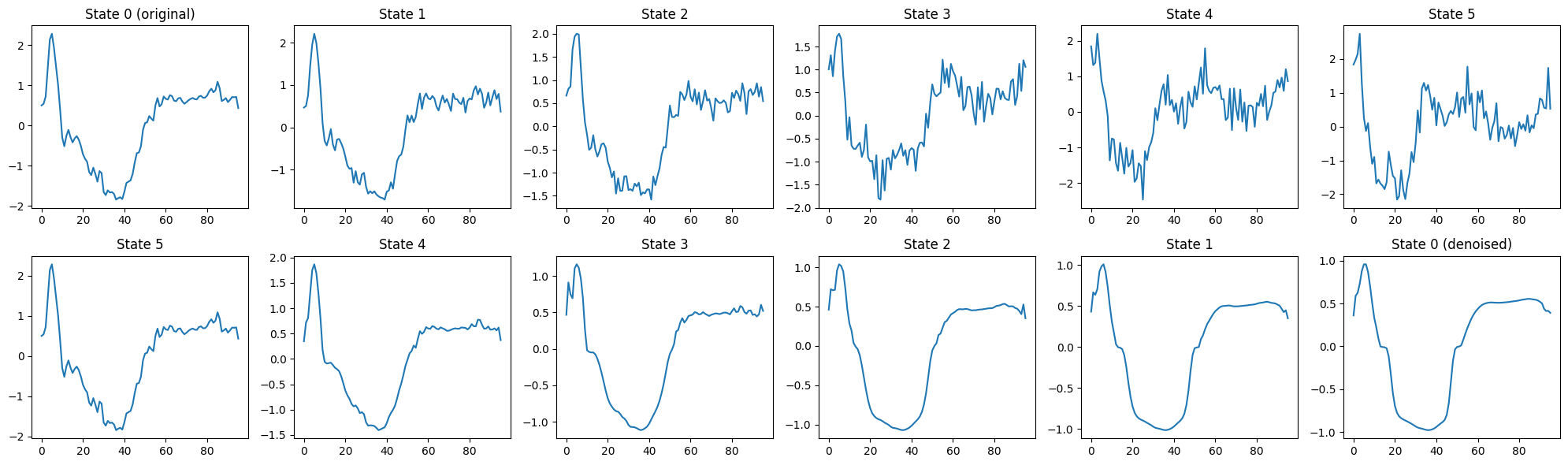} 
\caption{A sample diffusion process between two randomly selected samples within a class. The top row is the forward process in which we transform one sample from a class to another sample from the same class while adding Gaussian noise.}
\label{fig: sample_diffusion_within}
\end{figure*}

\subsection{Problem Definition}
\label{ss: definition}

Given a set of univariate time series $X=\left\{x_1, x_2, \ldots, x_N\right\}$, where each $x_i \in \mathbb{R}^M$ and has a corresponding binary label $y_i \in\{0,1\}$, our goal is to enhance the performance of a given deep learning model under multimodal, similar and noisy data distributions. Formally, the problem can be defined as:

\textbf{Multimodality}: The conditional distributions $\mathbb{P}(x \mid y=0)$ and $\mathbb{P}(x \mid y=1)$ can be expressed as mixtures of multiple component distributions, for example:
$$
\mathbb{P}(x \mid y=0)=\sum_{k=1}^{K_0} \pi_{0, k} \mathbb{P}_k(x \mid y=0)
$$
where $\pi_{0,k}$ is the mixture weight for the $k$-th component in the distributions for $y=0$, and $K_0$ is the number of components in each mixture.

\textbf{Similarity}: The similarity between the distributions $\mathbb{P}(x \mid y=0)$ and $\mathbb{P}(x \mid y=1)$ can be quantified using a distance measure $\mathcal{D}$, and we have:

$$
\mathcal{D}(\mathbb{P}(x \mid y=0) \| \mathbb{P}(x \mid y=1)) \text { is small.}
$$

\textbf{Noise}: Each time series $x$ is assumed to be a composition of a true underlying signal $s$ and additive noise $\eta$, such that:

$$
x=s+\eta
$$
where $s \in \mathbb{R}^M$ is the true signal and $\eta \in \mathbb{R}^M$ is the noise component.

\subsection{Architecture}

Figure \ref{fig:cdnet} provides an overview of the architecture. The proposed model consists of two phases: pre-training and fine-tuning. We explain the process starting with an example of Class 0.

During pre-training, to generate positive samples of Class 0, we construct a forward process by gradually linearly combining one sample with another within Class 0 and adding Gaussian noise. Then, we use CNNs to define the reverse process while minimizing the reconstruction error. For each sample in Class 0, we apply the trained reverse process and obtain one sample for each state; these states are positive samples for the given sample in Class 0. To generate negative samples, we follow the same procedure, except we linearly combine one sample from Class 0 with another sample from Class 1. This approach helps us learn the transition between samples from different classes, enabling us to better distinguish patterns when samples from the two labels are similar. We use the generated positive and negative sample pairs to minimize the loss. In the fine-tuning phase, we fix the weights learned from pre-training and fine-tune the last layer using cross-entropy loss. We repeat the same process for Class 1, resulting in a total of four diffusion processes.

\textbf{The feasibility of CNNs}. We justify the use of 1D convolutional neural networks (CNNs) to approximate the reverse diffusion process based on theoretical guarantees. Specifically, we show that CNNs can recover the denoising mapping $x^t \mapsto x^{t-1}$ under mild regularity assumptions. In Lemma \ref{lemma: 1dcnnapproximation} (Appendix), we assume that the clean time series signals are sampled from a compact subset of the Sobolev space $W^{1,2}([0,1])$, meaning they have bounded first derivatives and finite energy. Given that each noisy sample $x^t$ arises from a smooth affine transformation of $x^{t-1}$, plus additive Gaussian noise, the reverse map is Lipschitz continuous. Then, by the universal approximation theorem for CNNs over compact domains in Sobolev-type spaces \cite{yarotsky2017error, zhou2020universality}, there exists a 1D CNN that approximates the reverse transition $x^t \mapsto x^{t-1}$ to arbitrary precision in $\ell_2$ norm. This confirms that learning the reverse transitions via CNNs is not only empirically effective but also theoretically grounded.

Moreover, Lemma \ref{lemma:cnnconvergence} further establishes that CNN-based reverse mappings trained on interpolated and noisy forward samples can recover the support of the original multimodal data distribution. This indicates that CNN-based diffusion enables mode coverage, which is crucial for modeling the complex intra-class variability common in multimodal and noisy time series. Therefore, CNNs serve both as powerful denoisers and flexible generators of representative samples across modes.

\begin{lemma}[CNN-Based Reverse Diffusion Enables Mode Coverage]
\label{lemma:cnnconvergence}

Let $\mathbb{P}(x \mid y)$ be a class-conditional multimodal distribution over $x \in \mathbb{R}^M$, with $K$ modes:
\[
\mathbb{P}(x \mid y) = \sum_{k=1}^K \pi_k^{(y)} \mathbb{P}_k^{(y)}(x),
\]
where each $\mathbb{P}_k^{(y)}$ has disjoint support and $\pi_k^{(y)} > 0$. Define a stochastic forward process:
\[
x^t = \alpha x + (1 - \alpha) x' + \sqrt{\beta^t} \, \epsilon, \quad \epsilon \sim \mathcal{N}(0, \sigma^2 I),
\]
where $x, x' \sim \mathbb{P}(x \mid y)$ and $\alpha \in (0,1)$. Let $f_\theta$ be a 1D CNN trained to minimize the expected reconstruction loss:
\[
\min_\theta \mathbb{E}_{x, x'} \mathbb{E}_\epsilon \left[ \| f_\theta(x^t) - x \|_2^2 \right].
\]
Then under standard assumptions on the optimizer (e.g., SGD with diminishing step size) and sufficient model capacity, the learned distribution
\[
\hat{\mathbb{P}}_\theta(x \mid y) := f_\theta(x^t), \quad x^t \sim q_t(x^t \mid y)
\]
has support over $\bigcup_{k=1}^K \text{supp}(\mathbb{P}_k^{(y)})$.
\end{lemma}

Compared to diffusion in high-dimensional domains like images, the reverse mapping in 1D time series is structurally simpler and less sensitive to perturbations. Thus, the instability often observed when approximating reverse transitions in vision tasks may not hinder performance here; in fact, the mild randomness from CNN-based transitions may introduce useful diversity in generated samples for contrastive learning.

\begin{itemize}
    \item Within-class forward process:
$$
x_{i}^t=\sqrt{1-\beta^t} x_{i}^{t-1}+\left(1-\sqrt{1-\beta^t}\right) x_{j}^{0}+\sqrt{\beta^t} \epsilon
$$
for $x_i$ and $x_j$ from the same class. $x_{j}^{0}$ is a randomly selected raw sample.
    \item Across-class forward process:
$$
x_{i}^{t}=\sqrt{1-\beta^t} x_{i}^{t-1}+\left(1-\sqrt{1-\beta^t}\right) x_{c}^{0}+\sqrt{\beta^t} \epsilon
$$
for $x_i$ and $x_c$ from different classes.
where $\epsilon \sim \mathcal{N}(0, \sigma)$.
\end{itemize}

$\beta_{t}$ is the parameter controlling the rate of changes of diffusion process and $t \in \{1,...,T\}$ is the number of state steps.

The forward process is similar to Han et al.'s work except transitions are between different instances. However, the use CNNs denoted as $f_{\theta_t}$ to model the transitions in the reverse process as follows:

$$
x_{i}^{t-1}=f^t_{\theta_{t}} \left(x_{i}^{t}\right),
$$

where four sets of reverse CNNs need to be trained independently without parameter sharing.

Lastly, we train each reverse CNN $f^t_{\theta_t}$ to minimize the reconstruction error between $x^t$ and $x_i^{t-1}$, where $x^t$ is a noisy interpolated sample between $x_i$ and another within-class sample $x_j$. As shown in Lemma~\ref{lemma:denoising} (Appendix), this training objective corresponds to recovering a denoised estimate of $x_i^{t-1}$ in expectation, despite the presence of Gaussian noise and inter-sample mixing. Thus, each reverse CNN learns to model interpolation-aware denoising transitions.

Finally, given a time series $x_i$ from class $k$ (also known as anchor variables in contrastive learning), we select another time series $x_j$ from the same class $k$ and a time series $x_c$ from the other class $k^{\prime}$. We define the positive samples $x_i^{t+}$ of $x_i$ at state $t$ by:

$$
x_{i}^{t+}=f^1_{\theta_{1}} (f^2_{\theta_{2}}  (\cdots (f^t_{\theta_{t}}(x_{j}^{t} )))),
$$

and negative sample $x_i^{t-}$ by:

$$
x_{i}^{t-}=f^1_{\theta_{1}} (f^2_{\theta_{2}}  (\cdots (f^t_{\theta_{t}}(x_{c}^{t} )))).
$$

\begin{algorithm}
\caption{CDNet for Time Series Classification}
\begin{algorithmic}[1]
\STATE \textbf{Input:} Univariate time series $X$ with binary labels $Y$
\STATE \textbf{Output:} Predicted labels

\STATE \textbf{Pre-training Phase:}
\STATE Initialize $L_{\text{total}} \leftarrow 0$
\FOR{each sample pair $(x_i, x_j)$ within the same class}
    \STATE $x_{i}^t \leftarrow \sqrt{1 - \beta^t} x_{i}^{t-1} + (1 - \sqrt{1 - \beta^t}) x_{i}^0 + \sqrt{\beta^t} \epsilon$
    \STATE $x_{i}^{t-1} \leftarrow f_{\theta^t}(x_{i}^t)$ using CNNs
    \STATE Minimize reconstruction error
\ENDFOR
\FOR{each sample pair $(x_i, x_c)$ across different classes}
    \STATE $x_{i}^t \leftarrow \sqrt{1 - \beta^t} x_{i}^{t-1} + (1 - \sqrt{1 - \beta^t}) x_{c}^0 + \sqrt{\beta^t} \epsilon$
    \STATE $x_{i}^{t-1} \leftarrow f_{\theta_t}(x_{i}^t)$ using CNNs
    \STATE Minimize reconstruction error
\ENDFOR
\STATE Generate samples:
    \STATE Positive: $x_{i}^{t+}=f^1_{\theta_{1}} (f^2_{\theta_{2}}  (\cdots (f^t_{\theta_{t}}(x_{j}^{t} ))))$
    \STATE Negative: $x_{i}^{t-}=f^1_{\theta_{1}} (f^2_{\theta_{2}}  (\cdots (f^t_{\theta_{t}}(x_{c}^{t} ))))$

\STATE \textbf{Loss Calculation:}
\STATE Define composite loss:
\[
L_{\text{total}} = \omega_{\text{CE}} L_{\text{CE}} + \omega_{\text{SNN}} L_{\text{SNN}} + \omega_{\text{Triplet}} L_{\text{Triplet}}
\]

\STATE \textbf{Fine-tuning Phase:}
\STATE Fix pre-trained weights.
\STATE Fine-tune last layer with cross-entropy loss.

\STATE \textbf{Return} the trained model.

\STATE \textbf{End Procedure}
\end{algorithmic}
\end{algorithm}

\subsection{The Composite Loss}

Once positive and negative samples $\left\{x_i^{t+}\right\}_{t=1}^T$ and $\left\{x_i^{t-}\right\}_{t=1}^T$ of a sample $x_i$ are generated, we follow a conventional approach for pre-training and fine-tuning the classifier. During pre-training, instead of using fixed weighting parameters, we adopt an \textit{uncertainty-based weighting} strategy to dynamically learn the importance of each loss component \cite{kendall2018multi}. Specifically, we define the total loss as a combination of cross-entropy (CE) loss, soft nearest neighbour (SNN) loss, and triplet loss:

\begin{align*}
\mathcal{L}_{\text{total}} = 
& \ \frac{1}{2\sigma_{\mathrm{CE}}^2} \mathcal{L}_{\mathrm{CE}} 
+ \frac{1}{2\sigma_{\mathrm{SNN}}^2} \mathcal{L}_{\mathrm{SNN}} 
+ \frac{1}{2\sigma_{\mathrm{Triplet}}^2} \mathcal{L}_{\mathrm{Triplet}} \\
& + \log \sigma_{\mathrm{CE}} + \log \sigma_{\mathrm{SNN}} + \log \sigma_{\mathrm{Triplet}}
\end{align*}

Here, $\sigma_{\mathrm{CE}}, \sigma_{\mathrm{SNN}}, \sigma_{\mathrm{Triplet}}$ are learnable parameters representing task-dependent uncertainty. The loss terms with higher predictive uncertainty (larger $\sigma$) are down-weighted, while the $\log \sigma$ terms act as regularizers to prevent trivial solutions. This formulation allows the model to adaptively adjust the weight of each objective throughout training based on task difficulty and noise.

Given a sample, CDNet generates a sequence of negative samples using the diffusion process from samples across classes. Positive samples are generated using the diffusion process from samples within classes. Each triplet at each state is used for the triplet loss, and the original sample along with its neighbours is used for soft nearest neighbour loss and cross-entropy loss.

Specifically, 

$$
\mathcal{L}_{\text {Triplet }}=\sum_{i=i}^{T} \max \left(0,\left\|x_i-x_i^{t+}\right\|^2-\left\|x_i-x_i^{t-}\right\|^2+\alpha\right),
$$

where $\alpha$ is the margin.

The SNN loss is defined as follows:
\begin{align*}
& \mathcal{L}_{\mathrm{SNN}}\\
& =-\frac{1}{N} \sum_{i=1}^N \log \left(\frac{\exp \left(\frac{E_i \cdot E_{y_i}}{\tau}\right)}{\sum_{j=1}^N \exp \left(\frac{E_i E_j}{\tau}\right) \cdot \operatorname{mask}(i, j)+\epsilon}+\epsilon\right)
\end{align*}

where a small constant $\epsilon$ is added to avoid division by zero. $E_i$ is the embedding of the anchor sample $i$. $E_{y_i}$ is the embedding of the positive sample corresponding to the label $y_i$ passing through the classifier without the final layer. The temperature parameter $\tau$ scales the similarity scores, and $\operatorname{mask}(i, j)$ is a binary mask that excludes self-similarity.

Finally, we have CE loss defined as follows:
$$
\mathcal{L}_{\mathrm{CE}}=-\frac{1}{N} \sum_{i=1}^N  y_{i} \log \left(p_{i}\right).
$$

Finally, we fine-tune the model by freezing the weights trained in the previous layers, except for the last layers.

\section{Experiments}
In this section, we conduct a simulation study and evaluate the proposed model on the UCR Archive. We compare our model with SOTA TSC algorithms and variants of algorithms based on different loss functions. All experiments were implemented on a machine powered by an Intel i9-14900K processor with 64GB of memory and an NVIDIA RTX 4090 GPU with 24GB of memory. Python 3.9.16 and PyTorch 1.13.1 were utilized to construct and train our model. To demonstrate the the proposed CDNet will significantly improve the performance, we apply CDNet on top of InceptionTime (IT) \cite{ismail2020inceptiontime} and 1DCNN \cite{zhao2017convolutional}, and LSTM\_FCN \cite{karim2019multivariate}. We selected these three models as they are SOTA deep learning models and have relatively simple structures.

\subsection{UCR Archive}
\subsubsection{Variants Comparisons}
As aforementioned, the proposed architecture employs a composite loss of soft nearest neighbours loss, cross-entropy loss, and triplet loss. In this section, we will demonstrate the effectiveness of this choice of loss function through multiple tests on the UCR Archive. We compare the CDNet variants for both IT and 1DCNN. The detailed results could be found in Appendix A.1.

\begin{figure}[!ht]
\centering
\includegraphics[width=0.5\textwidth]{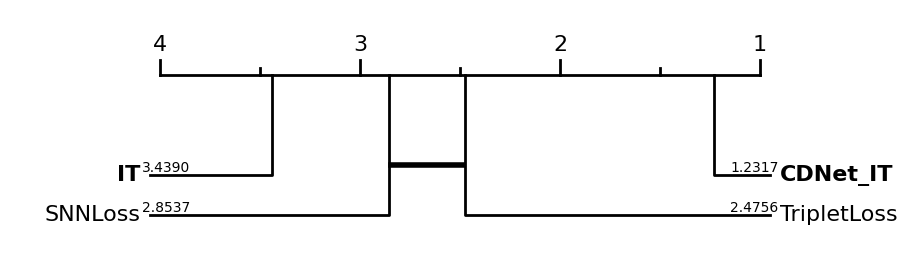}
\caption{Critical Difference (CD) diagram comparing CDNet variants on top of InceptionTime across all UCR Archive datasets. Lower ranks indicate better average performance. Connected models are not significantly different (p > 0.05, Friedman test).}
\label{img/fig:variant_it}
\end{figure}

\begin{figure}[!ht]
\centering
\includegraphics[width=0.5\textwidth]{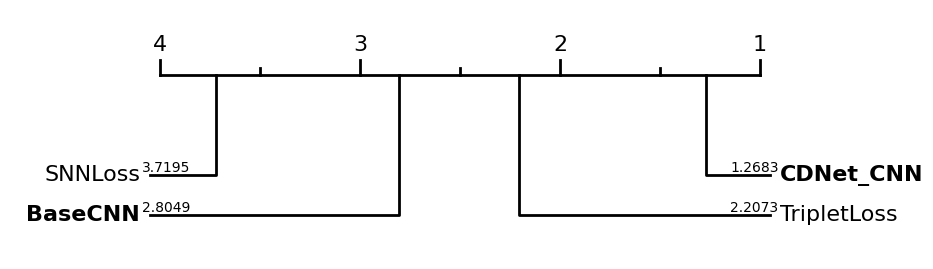}
\caption{
CD diagram comparing CDNet variants on top of 1DCNN across all binary UCR Archive datasets. 
}
\label{fig:variant_cnn}
\end{figure}

\begin{figure}[!ht]
\centering
\includegraphics[width=0.5\textwidth]{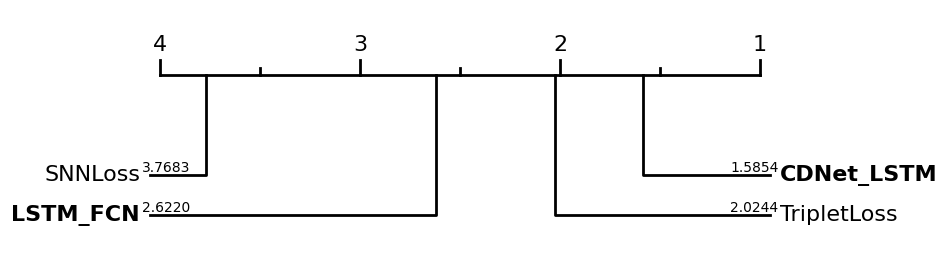}
\caption{CD diagram of CDNet variants with LSTM\_FCN on the UCR Archive.}
\label{fig:variant_lstm}
\end{figure}

The CD diagram for IT reveals that CDNet can significantly improve accuracies, as indicated by Figure \ref{img/fig:variant_it}. Similarly, the CD diagrams in Figure \ref{fig:variant_cnn} and Figure \ref{fig:variant_lstm} show that CDNet on top of 1DCNN significantly outperforms other choices of loss functions. Notably, the results obtained from the triplet loss ranked second best among all the choices of losses. This further demonstrates that the proposed mechanism can increase accuracies, as triplet loss directly uses the positive and negative samples generated from our diffusion process. In contrast, SNN loss often has the worst performance. One underlying reason could be that traditional distance metrics, such as cosine similarity used in SNN loss, cannot capture temporal dependencies effectively.

\subsubsection{SOTA Comparisons}

To ensure a fair evaluation of the improvement that CDNet brings to a particular model among SOTA models, we compare the performance of our proposed model using the UCR Archive with binary classification problems. We evaluate the proposed model against variants based on different loss functions, as well as with SOTA time series classification algorithms, including ShapeDTW (SDTW) \cite{zhao2018shapedtw}, Catch22 \cite{lubba2019catch22}, cBOSS \cite{middlehurst2019scalable}, WEASEL \cite{schafer2017fast}, HC2 \cite{middlehurst2021hive}, MultiRocket (MR) \cite{tan2022multirocket}, STSF \cite{cabello2020fast}, and FreshPRINCE (FP) \cite{middlehurst2022freshprince}. The detailed results could be found in Appendix A.1.

Figure~\ref{img/fig:variant_it} presents a Critical Difference (CD) diagram comparing the performance of InceptionTime (IT) and its CDNet-enhanced variants across all binary datasets in the UCR Archive. The diagram is based on Friedman’s test followed by the Nemenyi post-hoc analysis, a standard non-parametric approach in time series classification for comparing multiple models over multiple datasets. In this visualization, models are ranked by average accuracy rank (lower is better), and horizontal bars connect models that are not statistically significantly different at a 5\% significance level.

Notably, applying CDNet to IT leads to a substantial improvement in performance. The original IT model, while already competitive with cBOSS and Catch22, and close to MultiRocket and HC2, is outperformed by the CDNet-enhanced variant, which achieves the best average rank among all models. The lack of connection between the CDNet-enhanced IT and the other models indicates that the improvement is statistically significant.

This result highlights that CDNet can serve as a plug-in augmentation framework capable of consistently improving a strong baseline across diverse time series domains. Furthermore, the statistically significant gain underlines that CDNet does more than introduce random variance; it systematically enhances representation learning by generating informative contrastive pairs through its reverse diffusion mechanism. We further support this by providing CD diagrams across multiple backbones and by planning to visualize diffusion trajectories in future work to improve model interpretability under ambiguous and noisy conditions.

\begin{figure}[!ht]
\centering
\includegraphics[width=0.5\textwidth]{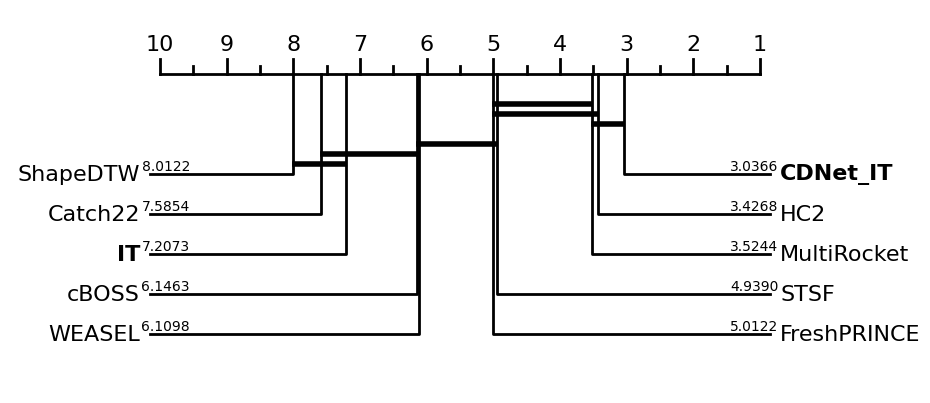}
\caption{CD diagram of CDNet on top of InceptionTime and SOTA models on the UCR Archive.}
\label{fig: sota_it}
\end{figure}

Similarly, Figure \ref{fig:variant_cnn} presents a CD diagram of CDNet on top of 1DCNN based on Friedman's test. It shows that the baseline 1DCNN (BaseCNN) has the lowest average rank among the selected SOTA models. However, when the proposed mechanism is applied to 1DCNN, the performance significantly improves, placing it among the top-performing SOTA models.

\begin{figure}[!ht]
\centering
\includegraphics[width=0.5\textwidth]{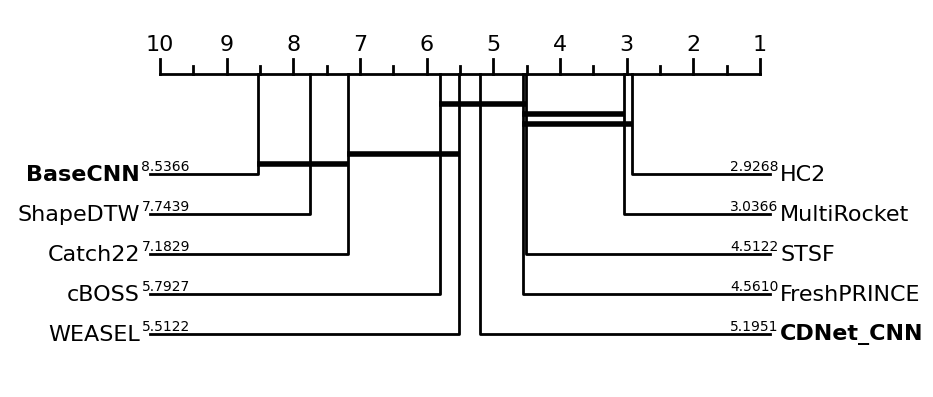}
\caption{CD diagram of CDNet on top of 1DCNN and SOTA models on the UCR Archive.}
\label{fig: sota_cnn}
\end{figure}

Lastly, Figure \ref{fig:variant_lstm} shows the improvement in performance of LSTM\_FCN. Although CDNet on top of LSTM\_FCN significantly outperforms the baseline LSTM\_FCN, the improvement is not as substantial as in the other two examples among SOTA models. One reason could be that LSTM\_FCN itself is not as effective as other SOTA models, limiting the potential for improvement. Another reason could be the interaction between the original model and the proposed mechanism of generating sample pairs.

All three examples demonstrate the scalability and effectiveness of CDNet. However, the most significant performance improvements among SOTA models are shown with IT and 1DCNN. One underlying reason for this significant boost in accuracy could be the interaction between the original classifiers and the CDNet mechanism. Both IT and 1DCNN use convolutional operations, which are originally designed to capture spatial dependencies but might not be as effective for capturing temporal dependencies. During our loss minimization process, we generate samples starting from the same anchor, as indicated by Figure \ref{fig: sample_diffusion_within}. This approach allows IT and 1DCNN to focus on capturing only local differences, thereby enhancing performance. This further explains the motivation behind the proposed idea.

\begin{figure}[!ht]
\centering
\includegraphics[width=0.5\textwidth]{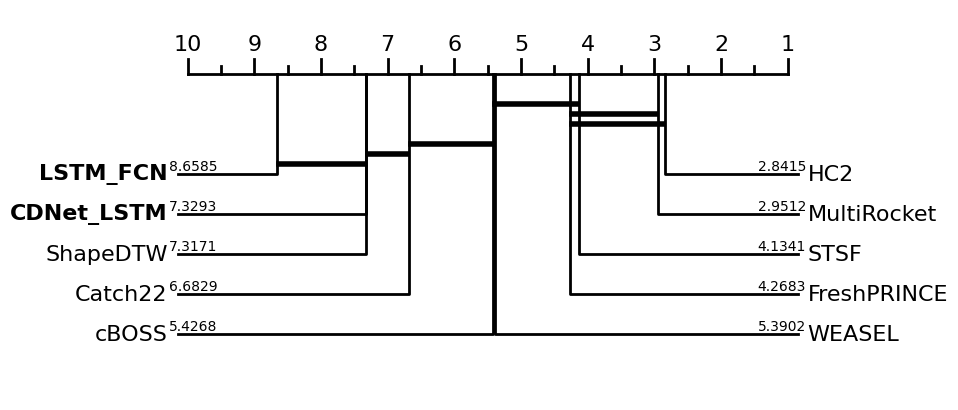}
\caption{CD diagram of CDNet on top of LSTM\_FCN and SOTA models on the UCR Archive.}
\label{fig: sota_lstm}
\end{figure}

\subsection{Simulation Study}
\subsubsection{Data Generation}
Although the targeted data characteristics commonly exist in the UCR Archive, we do not have an objective evaluation of the level of multimodality in each dataset. To study how each factor impacts the improvement of the accuracy of CDNet, we conducted a simulation study using sinusoidal curves. Specifically, we first define a base pattern:

$$
P(t)=A \cdot \sin (2 \pi f t+\phi)
$$
where $A$ is the amplitude, $f$ is the frequency, $\phi$ is the phase shift, and $t$ is the time index. The parameters are randomly chosen from specific ranges to introduce variability and simulate different data characteristics. Then we combine base patterns to generate complex patterns.

Given $k$ selected patterns $P_{i_1}(t), P_{i_2}(t), \ldots, P_{i_k}(t)$ and weights $w_1, w_2, \ldots, w_k$, the combined pattern $C(t)$ is:
$$
C(t)=\sum_{j=1}^k w_j \cdot P_{i_j}(t - \delta) + \mathcal{N}\left(0, \sigma\right)
$$
where $k$ is the number of selected patterns, $w_j$ are the weights drawn from a Dirichlet distribution, and $\delta$ is a random phase shift. The term $\mathcal{N}(0, \sigma^2)$ represents Gaussian noise with mean 0 and standard deviation $\sigma$, added to introduce variability and simulate noise in the data.

Finally, we generate two sets of patterns, one for each class, using different ranges for frequency, amplitude, and phase shift. We control the level of data characteristics by manipulating these parameters. For example, a less multimodal time series is sampled from smaller parameter ranges, resulting in a smaller variation of samples. All simulations and evaluations are repeated three times. The detailed process can be found in Appendix A.2.

\subsubsection{Data Characteristics Relationships}
\begin{figure}[!ht]
\centering
\includegraphics[width=0.46\textwidth]{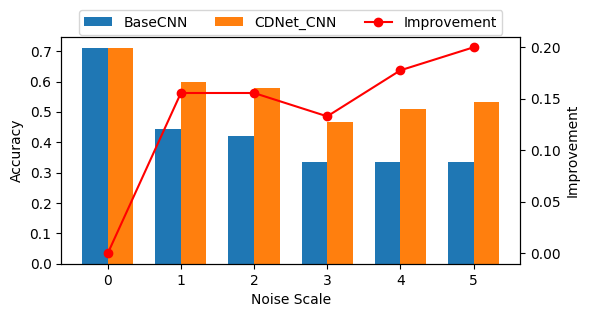}
\caption{Average accuracy of BaseCNN and CDNet\_CNN and improvement wrt noise level. A higher noise level means the sample is noisier.}
\label{fig: simu_noise}
\end{figure}

\begin{figure}[!ht]
\centering
\includegraphics[width=0.46\textwidth]{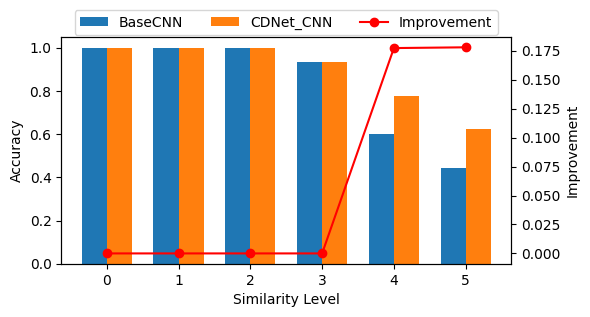}
\caption{Average accuracy of BaseCNN and CDNet\_CNN and improvement wrt class similarity level. A higher similarity level means samples from the two classes are more similar.}
\label{fig: simu_similar}
\end{figure}

\begin{figure}[!ht]
\centering
\includegraphics[width=0.46\textwidth]{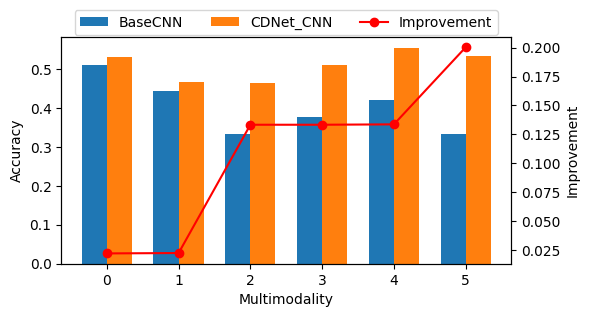}
\caption{Average accuracy of BaseCNN and CDNet\_CNN and improvement wrt class multimodal level. A higher multimodal level means samples are more multimodal.}
\label{fig: simu_multimodal}
\end{figure}

We explore the relationship between the performance of CDNet on top of 1DCNN concerning noise scale, similarity level, and multimodality. We choose 1DCNN since it is simple. Figure \ref{fig: simu_noise} shows that as the simulated samples become noisier, the performance of both CDNet\_CNN and BaseCNN decreases, but the improvement in accuracies increases. Additionally, CDNet exhibits smaller changes in accuracies with increasing noise scale. Similar patterns are observed for class similarity level and multimodality, as shown in Figure \ref{fig: simu_similar} and Figure \ref{fig: simu_multimodal}, respectively. Therefore, our model demonstrates robustness in the presence of noise, class similarity, and multimodality.

\section{Conclusion}

In this work, we proposed a novel algorithm, CDNet, for univariate binary TSC, leveraging principles of diffusion models and contrastive learning to enhance the performance of SOTA deep learning classifiers. We generate positive and negative sample pairs through a CNN-based diffusion process by considering transitions between samples within and across classes. During pre-training, we minimize a composite of contrastive loss, and we fine-tune the model by freezing the weights. This architecture effectively learns the embedding space for time series with similar and multimodal label distributions, especially under noisy distributions, significantly improving the performance of a given deep learning model. We have shown that CDNet on top of IT, 1DCNN, and LSTM\_FCN significantly outperforms the baselines, with IT outperforming all the SOTA models when complemented with CDNet.

However, there are areas for future improvement. Currently, CDNet is suitable only for univariate binary TSC. For time series with more dimensions and labels, structural changes are needed. Additionally, CDNet relies on knowing the detailed architecture of a given classifier. Future research could focus on making the mechanism effective without requiring knowledge of the specific model architecture. While many SOTA models like MultiRocket and HC2 perform implicit augmentation via randomized convolutions or ensembling, we agree explicit comparisons to contrastive-augmentation baselines would strengthen our contributions.

\begin{ack}
We thank the UCR Time Series Classification Archive team \cite{dau2019ucr} for providing a comprehensive and well-curated benchmark dataset that has been instrumental in the development and evaluation of our work. 
\end{ack}



\newpage
\appendix
\onecolumn
\section{Appendix}

\subsection{Theoretical Analysis}

\noindent\textbf{Proof of Lemma~\ref{lemma:cnnconvergence}}.
\begin{proof}
Define the forward sample distribution $q_t(x^t \mid y)$ induced by sampling $x, x' \sim \mathbb{P}(x \mid y)$ and $\epsilon \sim \mathcal{N}(0, \sigma^2 I)$:
\[
x^t = \alpha x + (1 - \alpha) x' + \sqrt{\beta^t} \, \epsilon.
\]
As $x$ and $x'$ vary across different modes $\mathbb{P}_k^{(y)}$, and $\epsilon$ adds full-rank Gaussian noise, the support of $q_t(x^t \mid y)$ becomes a connected region covering the convex hull of all mode supports. That is,
\[
\text{supp}(q_t(x^t \mid y)) \supseteq \text{conv}\left( \bigcup_{k=1}^K \text{supp}(\mathbb{P}_k^{(y)}) \right).
\]

Now, define the loss function:
\[
\mathcal{L}(\theta) = \mathbb{E}_{x, x'} \mathbb{E}_\epsilon \left[ \| f_\theta(x^t) - x \|_2^2 \right].
\]
This is a standard stochastic reconstruction objective. Under sufficient capacity (i.e., $f_\theta$ is a universal approximator on compact domains \cite{yarotsky2017error, zhou2020universality}), and convexity of $\| \cdot \|_2^2$ loss in $f_\theta$, the optimization via SGD converges to a stationary point:
\[
\nabla_\theta \mathcal{L}(\theta) = 0.
\]

At convergence, $f_\theta$ maps each $x^t$ toward its corresponding $x$, which comes from a particular mode $\mathbb{P}_k^{(y)}$. Since $q_t(x^t \mid y)$ includes interpolated/noisy mixtures between modes, $f_\theta$ learns to map different neighborhoods in $x^t$ space to different $x$ from different modes.

Therefore, the pushforward distribution $\hat{\mathbb{P}}_\theta(x \mid y)$ induced by sampling $x^t \sim q_t(x^t \mid y)$ and applying $f_\theta$ satisfies:
\[
\text{supp}(\hat{\mathbb{P}}_\theta(x \mid y)) \supseteq \bigcup_{k=1}^K \text{supp}(\mathbb{P}_k^{(y)}).
\]
\end{proof}

\begin{lemma}[1D CNNs Approximate Reverse Diffusion Transitions]
\label{lemma: 1dcnnapproximation}

Let $\mathcal{F} \subset W^{1,2}([0,1])$ be a class of univariate time series functions with bounded first derivatives and compact support. Suppose each observed time series $x^t \in \mathbb{R}^M$ is a discretization of some $f \in \mathcal{F}$, sampled at $M$ equally spaced time points. Let the forward diffusion process at time $t$ be defined as:
\[
x^t = \sqrt{1 - \beta^t} \, x^{t-1} + \left(1 - \sqrt{1 - \beta^t} \right) x_j^0 + \sqrt{\beta^t} \, \epsilon^t,
\]
where $x^{t-1}, x_j^0 \in \mathbb{R}^M$, $\epsilon^t \sim \mathcal{N}(0, \sigma^2 I)$, and $\beta^t \leq \beta_{\max} \ll 1$. Then for any $\varepsilon > 0$, there exists a 1D convolutional neural network $f_\theta: \mathbb{R}^M \to \mathbb{R}^M$ such that
\[
\sup_{x^t \in \mathcal{F}_M} \left\| f_\theta(x^t) - x^{t-1} \right\|_2 \leq \varepsilon,
\]
where $\mathcal{F}_M$ denotes the set of discretized time series derived from $\mathcal{F}$.

\begin{proof}
We aim to show that for any $\varepsilon > 0$, there exists a 1D convolutional neural network $f_\theta: \mathbb{R}^M \to \mathbb{R}^M$ such that $\sup_{x^t \in \mathcal{F}_M} \|f_\theta(x^t) - x^{t-1}\|_2 \leq \varepsilon$, where $x^t$ is the result of the forward diffusion process and $x^{t-1}$ is the desired reconstruction.

The forward process is given by
\[
x^t = \sqrt{1 - \beta^t} \, x^{t-1} + \left(1 - \sqrt{1 - \beta^t} \right) x_j^0 + \sqrt{\beta^t} \, \epsilon^t,
\]
where $\epsilon^t \sim \mathcal{N}(0, \sigma^2 I)$, and $x_j^0$ is a second sample from the same class. Solving for $x^{t-1}$ yields the reverse mapping
\[
x^{t-1} = \frac{1}{\sqrt{1 - \beta^t}} \left( x^t - \left(1 - \sqrt{1 - \beta^t} \right) x_j^0 - \sqrt{\beta^t} \, \epsilon^t \right),
\]
which we denote by $g(x^t) = x^{t-1}$. This function is affine in $x^t$, and therefore Lipschitz continuous. In particular, for any $x_1^t, x_2^t \in \mathcal{F}_M$, we have
\[
\|g(x_1^t) - g(x_2^t)\|_2 = \frac{1}{\sqrt{1 - \beta^t}} \|x_1^t - x_2^t\|_2,
\]
which implies that $g$ is $L$-Lipschitz with $L = \frac{1}{\sqrt{1 - \beta^t}} > 0$.

Assume that the clean time series functions belong to a compact subset $\mathcal{F} \subset W^{1,2}([0,1])$, and let $\mathcal{F}_M$ denote their discretizations over $M$ evenly spaced time points. Since $W^{1,2}([0,1])$ embeds continuously into $C([0,1])$ and $\mathcal{F}$ is compact in the Sobolev norm, the discretized set $\mathcal{F}_M$ is compact in $\mathbb{R}^M$.

By the universal approximation theorem for convolutional neural networks on Sobolev-type function spaces \cite{yarotsky2017error, zhou2020universality}, any Lipschitz continuous function defined on a compact subset of $\mathbb{R}^M$ can be approximated arbitrarily well in $\ell^2$ norm by a CNN with ReLU activations and sufficient depth, width, and receptive field. Therefore, for any $\varepsilon > 0$, there exists a CNN $f_\theta$ such that
\[
\sup_{x^t \in \mathcal{F}_M} \|f_\theta(x^t) - g(x^t)\|_2 = \sup_{x^t \in \mathcal{F}_M} \|f_\theta(x^t) - x^{t-1}\|_2 \leq \varepsilon.
\]
\end{proof}

\end{lemma}

Lemma~\ref{lemma: 1dcnnapproximation} holds for both within-class and across-class forward diffusion processes, as the interpolation term $x_j^0$ may be drawn from either the same or a different class. The approximation capacity of CNNs applies regardless of this class relationship, since the structural form of the forward process remains unchanged.

\begin{lemma}[Reverse Diffusion Learns Interpolated Denoising Transitions]
\label{lemma:denoising}

Let $x_i, x_j \in \mathbb{R}^M$ be two time series samples from the same class. Define the forward diffusion process at step $t$ as:
\[
x^t = \sqrt{1 - \beta^t} \, x_i^{t-1} + (1 - \sqrt{1 - \beta^t}) \, x_j^0 + \sqrt{\beta^t} \, \varepsilon^t, \quad \varepsilon^t \sim \mathcal{N}(0, \sigma^2 I),
\]
where $x_i^{t-1} = s_i^{t-1} + \eta$, with $s_i^{t-1}$ denoting the clean signal and $\eta$ zero-mean Gaussian noise independent of $\varepsilon^t$.

Let $f^t_{\theta_t}$ be a CNN trained to minimize the expected reconstruction loss:
\[
\min_{\theta_t} \mathbb{E}_{x_i, x_j, \varepsilon^t} \left\| f^t_{\theta_t}(x^t) - x_i^{t-1} \right\|_2^2.
\]
Then the optimal solution satisfies:
\[
f^t_{\theta_t}(x^t) = \mathbb{E}[x_i^{t-1} \mid x^t],
\]
and the expected squared error with respect to the clean signal is minimized up to the variance of the additive noise:
\[
\mathbb{E} \left\| f^t_{\theta_t}(x^t) - s_i^{t-1} \right\|_2^2 = \min + \mathrm{Var}(\eta).
\]

\begin{proof}
We are given a stochastic forward process:
\[
x^t = \sqrt{1 - \beta^t} \, x_i^{t-1} + (1 - \sqrt{1 - \beta^t}) \, x_j^0 + \sqrt{\beta^t} \, \varepsilon^t,
\]
and aim to learn a function $f^t_{\theta_t}$ that minimizes:
\[
\mathbb{E}_{x^t} \left\| f^t_{\theta_t}(x^t) - x_i^{t-1} \right\|_2^2.
\]
This is a standard mean squared error (MSE) regression setup. The theory of MMSE estimators tells us that the minimizer of this objective is:
\[
f^t_{\theta_t}(x^t) = \mathbb{E}[x_i^{t-1} \mid x^t].
\]

Now, assume $x_i^{t-1} = s_i^{t-1} + \eta$ with $\eta$ independent of $x^t$ and zero-mean. Then:
\[
\mathbb{E} \left\| f^t_{\theta_t}(x^t) - s_i^{t-1} \right\|_2^2 
= \mathbb{E} \left\| f^t_{\theta_t}(x^t) - x_i^{t-1} + \eta \right\|_2^2
= \mathbb{E} \left\| f^t_{\theta_t}(x^t) - x_i^{t-1} \right\|_2^2 + \mathbb{E} \left\| \eta \right\|_2^2,
\]
where we used independence and $\mathbb{E}[\eta] = 0$ to eliminate the cross-term.
Therefore, the expected squared error is minimized up to the irreducible variance introduced by the additive noise, that is,

$$
\mathbb{E}\left[\left\|f_\theta\left(x_t\right)-s_{t-1}\right\|^2\right]=\min _\theta \mathbb{E}\left[\left\|f_\theta\left(x_t\right)-x_{t-1}\right\|^2\right]+\operatorname{Var}(\eta). 
$$

This proves that the reverse diffusion step trained via CNN regression effectively performs interpolation-aware denoising in expectation.
\end{proof}

\end{lemma}

\begin{lemma}[Reverse Diffusion Distinguishes Across-Class Interpolations]
\label{lemma:acrossclass_denoising}

Let $x_i \in \mathbb{R}^M$ be a time series from class $y_i$ and $x_c \in \mathbb{R}^M$ be a time series from a different class $y_c \neq y_i$. Define the across-class forward diffusion process at step $t$ as:
\[
x^t = \sqrt{1 - \beta^t} \, x_i^{t-1} + (1 - \sqrt{1 - \beta^t}) \, x_c^0 + \sqrt{\beta^t} \, \varepsilon^t, \quad \varepsilon^t \sim \mathcal{N}(0, \sigma^2 I),
\]
where $x_i^{t-1} = s_i^{t-1} + \eta$ and $\eta$ is zero-mean Gaussian noise independent of $\varepsilon^t$.

Let $f^t_{\theta_t}$ be a CNN trained to minimize:
\[
\min_{\theta_t} \mathbb{E}_{x_i, x_c, \varepsilon^t} \left\| f^t_{\theta_t}(x^t) - x_i^{t-1} \right\|_2^2.
\]
Then the optimal solution satisfies:
\[
f^t_{\theta_t}(x^t) = \mathbb{E}[x_i^{t-1} \mid x^t],
\]
and the expected squared error to the clean signal satisfies:
\[
\mathbb{E} \left\| f^t_{\theta_t}(x^t) - s_i^{t-1} \right\|_2^2 = \min + \mathrm{Var}(\eta).
\]

\begin{proof}
This follows the same structure as Lemma~\ref{lemma:denoising}. The forward process includes interpolation between samples from different classes:
\[
x^t = \sqrt{1 - \beta^t} \, x_i^{t-1} + (1 - \sqrt{1 - \beta^t}) \, x_c^0 + \sqrt{\beta^t} \, \varepsilon^t.
\]
As before, $x^t$ is a noisy mixture, and we wish to reconstruct $x_i^{t-1}$. The MMSE estimator for minimizing:
\[
\mathbb{E} \left\| f^t_{\theta_t}(x^t) - x_i^{t-1} \right\|_2^2
\]
is:
\[
f^t_{\theta_t}(x^t) = \mathbb{E}[x_i^{t-1} \mid x^t].
\]

If $x_i^{t-1} = s_i^{t-1} + \eta$ and $\eta$ is independent of $x^t$, we have:
\[
\mathbb{E} \left\| f^t_{\theta_t}(x^t) - s_i^{t-1} \right\|_2^2 
= \mathbb{E} \left\| f^t_{\theta_t}(x^t) - x_i^{t-1} \right\|_2^2 + \mathrm{Var}(\eta).
\]

Thus, training on across-class interpolations allows the reverse CNN to learn discriminative denoising behavior by projecting mixed-class inputs back to the original class manifold.
\end{proof}

\end{lemma}

\subsection{Performance Evaluations}
In this section, we compare the enhanced performance of CDNet when applied to InceptionTime (IT), 1-dimensional convolutional neural network (1DCNN), and LSTM\_FCN as shown in Table \ref{tab: table1} (Appendix).

Table \ref{tab: table2} (Appendix) illustrates how CDNet improves the performance of IT, 1DCNN, and LSTM\_FCN among state-of-the-art models, including ShapeDTW (SDTW) \cite{zhao2018shapedtw}, Catch22 \cite{lubba2019catch22}, cBOSS \cite{middlehurst2019scalable}, WEASEL \cite{schafer2017fast}, HC2 \cite{middlehurst2021hive}, MultiRocket (MR) \cite{tan2022multirocket}, STSF \cite{cabello2020fast}, and FreshPRINCE (FP) \cite{middlehurst2022freshprince}. We follow the default parameter setting for each SOTA models.

\begin{table*}[ht]
\caption{Accuracy comparisons between CDNet models and baseline deep learning models on all binary UCR Archive datasets, including the change in accuracy for CDNet models compared to base models.}
\centering
\fontsize{9}{10}\selectfont 
\vskip 0.15in
\begin{tabular}{|l|l l l|l l l|l l l|}
\toprule
\multicolumn{1}{|c|}{} & \multicolumn{3}{c|}{\bf CDNet Models} & \multicolumn{3}{c|}{\bf Base Deep Learning Models} & \multicolumn{3}{c|}{\bf Change in Accuracy} \\
\midrule
\bf Problem & \bf CDNet\_IT & \bf CDNet\_CNN & \bf CDNet\_LSTM & \bf IT & \bf 1DCNN & \bf LSTM\_FCN & \bf $\delta$ IT & \bf $\delta$ CNN & \bf $\delta$ LSTM \\
\midrule
BF & 0.900 & 0.900 & \textbf{1.000} & 0.800 & 0.800 & 0.850 & 0.100 & 0.100 & 0.200 \\
BC & 0.950 & \textbf{1.000} & 0.994 & 0.800 & 0.750 & 0.550 & 0.150 & 0.200 & 0.194 \\
Chinatown & 0.983 & \textbf{1.000} & 0.994 & 0.977 & 0.750 & 0.834 & 0.006 & 0.023 & 0.017 \\
Coffee & \textbf{1.000} & \textbf{1.000} & 0.988 & 0.821 & \textbf{1.000} & 0.929 & 0.179 & 0.179 & 0.167 \\
Computers & 0.800 & 0.788 & \textbf{0.978} & 0.736 & 0.692 & 0.576 & 0.064 & 0.052 & 0.242 \\
DPOC & 0.793 & 0.754 & \textbf{0.977} & 0.768 & 0.743 & 0.743 & 0.025 & -0.014 & 0.209 \\
DLG & 0.920 & 0.783 & \textbf{0.964} & 0.732 & 0.754 & 0.688 & 0.188 & 0.051 & 0.232 \\
DLW & \textbf{0.978} & 0.971 & 0.950 & \textbf{0.964} & 0.891 & 0.942 & 0.014 & 0.007 & -0.014 \\
Earthquakes & 0.777 & 0.777 & \textbf{0.945} & 0.748 & 0.748 & 0.734 & 0.029 & 0.029 & 0.197 \\
ECG200 & \textbf{0.910} & \textbf{0.910} & 0.935 & 0.860 & 0.820 & 0.790 & 0.050 & 0.050 & 0.075 \\
ECGFD & \textbf{1.000} & 0.983 & 0.932 & 0.813 & 0.965 & 0.704 & 0.187 & 0.170 & 0.119 \\
FordA & \textbf{0.998} & 0.932 & 0.930 & 0.964 & 0.908 & 0.780 & 0.034 & -0.032 & -0.034 \\
FordB & \textbf{0.891} & 0.803 & 0.927 & 0.857 & 0.751 & 0.577 & 0.033 & -0.054 & 0.070 \\
FRT & \textbf{0.998} & \textbf{0.998} & 0.920 & 0.986 & 0.901 & 0.784 & 0.012 & 0.012 & -0.066 \\
FST & 0.801 & 0.861 & \textbf{0.906} & 0.701 & 0.737 & 0.707 & 0.100 & 0.160 & 0.205 \\
GP & \textbf{1.000} & 0.967 & 0.884 & 0.973 & 0.920 & 0.700 & 0.027 & -0.006 & -0.089 \\
GPAS & 0.994 & 0.984 & 0.880 & 0.965 & \textbf{0.972} & 0.883 & 0.029 & 0.019 & -0.085 \\
GPMVF & \textbf{1.000} & 0.991 & 0.856 & 0.889 & \textbf{0.978} & 0.968 & 0.111 & 0.022 & -0.033 \\
GPOVY & \textbf{1.000} & 0.927 & 0.850 & \textbf{1.000} & 0.841 & 0.902 & 0.000 & -0.073 & -0.150 \\
Ham & 0.781 & 0.743 & \textbf{0.840} & 0.657 & 0.762 & 0.705 & 0.124 & 0.086 & 0.183 \\
HO & \textbf{0.935} & \textbf{0.935} & 0.761 & 0.641 & 0.733 & 0.900 & 0.294 & 0.294 & 0.120 \\
Herring & 0.688 & 0.609 & \textbf{0.755} & 0.594 & 0.578 & 0.578 & 0.094 & 0.015 & 0.161 \\
HT & \textbf{0.967} & 0.950 & 0.752 & 0.950 & 0.594 & 0.731 & 0.017 & 0.000 & -0.198 \\
IPD & \textbf{0.969} & 0.861 & 0.752 & 0.951 & 0.950 & 0.953 & 0.018 & -0.090 & -0.199 \\
Lightning2 & \textbf{0.787} & 0.771 & 0.749 & 0.656 & 0.570 & 0.770 & 0.131 & 0.115 & 0.093 \\
MPOC & \textbf{0.873} & 0.801 & 0.739 & 0.570 & 0.800 & 0.732 & 0.303 & 0.231 & 0.169 \\
MS & \textbf{0.900} & 0.867 & 0.735 & 0.800 & 0.772 & 0.561 & 0.100 & 0.067 & -0.065 \\
POC & \textbf{0.853} & 0.807 & 0.705 & 0.772 & 0.856 & 0.669 & 0.081 & 0.035 & -0.067 \\
PC & \textbf{0.983} & 0.933 & 0.676 & 0.856 & 0.839 & 0.961 & 0.127 & 0.077 & -0.180 \\
PPOC & \textbf{0.904} & 0.876 & 0.628 & 0.839 & 0.653 & 0.749 & 0.065 & 0.037 & -0.211 \\
SHGCh2 & 0.853 & \textbf{0.855} & 0.623 & 0.653 & 0.500 & 0.752 & 0.200 & 0.202 & -0.030 \\
SS & \textbf{1.000} & \textbf{1.000} & 0.623 & 0.500 & 0.890 & 0.993 & 0.500 & 0.500 & 0.123 \\
Sony1 & 0.900 & \textbf{0.869} & 0.622 & 0.890 & 0.890 & 0.593 & 0.010 & -0.021 & -0.268 \\
Sony2 & 0.853 & \textbf{0.924} & 0.597 & 0.890 & 0.643 & 0.558 & -0.037 & 0.034 & -0.293 \\
Strawberry & \textbf{0.983} & 0.968 & 0.594 & 0.643 & 0.890 & 0.854 & 0.340 & 0.325 & -0.049 \\
ToeS1 & 0.904 & \textbf{0.939} & 0.587 & 0.890 & 0.700 & 0.654 & 0.014 & 0.049 & -0.303 \\
ToeS2 & \textbf{0.953} & 0.886 & 0.569 & 0.700 & \textbf{0.984} & 0.621 & 0.253 & 0.186 & -0.131 \\
TLECG & \textbf{0.998} & 0.985 & 0.500 & 0.984 & 0.981 & \textbf{0.993} & 0.014 & 0.001 & -0.484 \\
Wafer & \textbf{1.000} & 0.995 & \textbf{1.000} & 0.500 & 0.500 & 0.593 & 0.500 & 0.495 & 0.500 \\
Wine & 0.722 & \textbf{0.870} & 0.994 & 0.688 & 0.688 & 0.558 & 0.034 & 0.182 & 0.306 \\
WTC & 0.779 & 0.740 & \textbf{0.994} & 0.675 & 0.675 & 0.558 & 0.104 & 0.065 & 0.319 \\
\bottomrule
\end{tabular}
\label{tab: table1}
\end{table*}

\begin{table*}[ht]
\caption{Accuracy comparisons between CDNet models and SOTA non deep learning models on all binary UCR Archive datasets.}
\centering
\fontsize{9}{10}\selectfont 
\vskip 0.15in
\begin{tabular}{|l|l l l|l l l l l l l l|}
\toprule
\multicolumn{1}{|c|}{} & \multicolumn{3}{c|}{\bf CDNet Models} & \multicolumn{8}{c|}{\bf Non Deep Learning Models} \\
\midrule
\bf Problem & \bf IT & \bf 1DCNN & \bf LSTM\_FCN & Catch22 & MR & cBOSS & STSF & WEASEL & FP & HC2 & SDTW \\
\midrule
BF & 0.900 & 0.900 & \textbf{1.000} & 0.750 & 0.900 & \textbf{1.000} & 0.900 & 0.750 & 0.850 & 0.950 & 0.750 \\
BC & 0.950 & \textbf{1.000} & 0.994 & 0.900 & 0.900 & 0.950 & \textbf{1.000} & 0.900 & 0.950 & 0.900 & 0.550 \\
Chinatown & 0.983 & \textbf{1.000} & 0.994 & 0.886 & 0.977 & 0.936 & 0.968 & 0.948 & 0.977 & 0.983 & 0.962 \\
Coffee & \textbf{1.000} & \textbf{1.000} & 0.988 & \textbf{1.000} & \textbf{1.000} & \textbf{1.000} & \textbf{1.000} & \textbf{1.000} & \textbf{1.000} & \textbf{1.000} & \textbf{1.000} \\
Computers & 0.800 & 0.788 & \textbf{0.978} & 0.712 & 0.776 & 0.688 & 0.752 & 0.700 & 0.768 & 0.784 & 0.556 \\
DPOC & 0.793 & 0.754 & \textbf{0.977} & 0.783 & 0.794 & 0.725 & 0.790 & 0.790 & 0.786 & 0.750 & 0.721 \\
DLG & 0.920 & 0.783 & \textbf{0.964} & 0.739 & 0.826 & 0.688 & 0.855 & 0.594 & 0.899 & 0.899 & 0.891 \\
DLW & 0.978 & 0.971 & 0.950 & 0.942 & 0.978 & 0.971 & 0.978 & 0.920 & \textbf{0.986} & 0.978 & 0.978 \\
Earthquakes & 0.777 & 0.777 & \textbf{0.945} & 0.748 & 0.748 & 0.748 & 0.755 & 0.748 & 0.777 & 0.748 & 0.734 \\
ECG200 & \textbf{0.935} & \textbf{0.935} & 0.910 & 0.840 & \textbf{0.910} & 0.850 & 0.880 & 0.850 & 0.890 & 0.870 & 0.870 \\
ECGFD & \textbf{1.000} & 0.983 & 0.932 & 0.781 & \textbf{1.000} & \textbf{1.000} & \textbf{1.000} & 0.997 & \textbf{1.000} & \textbf{1.000} & 0.920 \\
FordA & \textbf{0.998} & 0.932 & 0.930 & 0.914 & 0.955 & 0.910 & 0.946 & 0.953 & \textbf{1.000} & 0.958 & 0.699 \\
FordB & \textbf{0.927} & 0.803 & 0.891 & 0.742 & 0.837 & 0.791 & 0.814 & 0.815 & 0.810 & 0.821 & 0.619 \\
FRT & \textbf{0.998} & \textbf{0.998} & 0.920 & 0.997 & \textbf{1.000} & 0.994 & \textbf{0.998} & 0.984 & 0.993 & 0.999 & 0.804 \\
FST & 0.801 & 0.861 & \textbf{0.906} & 0.950 & 0.994 & 0.971 & \textbf{0.998} & 0.970 & 0.991 & 0.999 & 0.676 \\
GP & \textbf{1.000} & 0.967 & 0.884 & 0.967 & \textbf{1.000} & \textbf{1.000} & 0.920 & \textbf{1.000} & 0.933 & \textbf{1.000} & 0.960 \\
GPAS & 0.994 & 0.984 & 0.880 & 0.981 & \textbf{1.000} & \textbf{1.000} & 0.975 & 0.994 & 0.978 & 0.997 & 0.984 \\
GPMVF & \textbf{1.000} & 0.991 & 0.856 & 0.994 & \textbf{1.000} & \textbf{1.000} & \textbf{1.000} & 0.991 & 0.984 & \textbf{1.000} & \textbf{1.000} \\
GPOVY & \textbf{1.000} & 0.927 & 0.850 & \textbf{1.000} & \textbf{1.000} & 0.997 & \textbf{1.000} & 0.987 & \textbf{1.000} & \textbf{1.000} & \textbf{1.000} \\
Ham & \textbf{0.840} & 0.743 & 0.781 & 0.610 & 0.733 & 0.676 & 0.733 & 0.714 & 0.714 & 0.762 & 0.600 \\
HO & \textbf{0.951} & 0.935 & 0.761 & 0.865 & \textbf{0.951} & 0.908 & 0.922 & 0.887 & 0.905 & 0.932 & 0.851 \\
Herring & \textbf{0.755} & 0.609 & 0.688 & 0.563 & \textbf{0.734} & 0.578 & 0.641 & 0.594 & 0.641 & 0.672 & 0.531 \\
HT & \textbf{0.967} & 0.950 & 0.752 & 0.941 & \textbf{0.983} & 0.975 & 0.882 & 0.950 & 0.924 & 0.966 & 0.681 \\
IPD & \textbf{0.970} & 0.861 & 0.752 & 0.896 & \textbf{0.970} & 0.910 & 0.967 & 0.945 & 0.899 & \textbf{0.970} & 0.965 \\
Lightning2 & 0.787 & 0.771 & 0.749 & 0.721 & 0.672 & 0.771 & 0.705 & 0.607 & 0.738 & 0.787 & \textbf{0.803} \\
MPOC & 0.873 & 0.801 & 0.739 & 0.777 & 0.856 & 0.784 & 0.828 & 0.777 & 0.866 & 0.849 & \textbf{0.873} \\
MS & \textbf{0.900} & 0.867 & 0.735 & 0.859 & \textbf{0.943} & 0.918 & 0.853 & 0.938 & 0.898 & 0.966 & 0.879 \\
POC & \textbf{0.853} & 0.807 & 0.705 & 0.791 & 0.850 & 0.768 & 0.839 & 0.790 & 0.838 & 0.832 & 0.760 \\
PC & \textbf{0.983} & 0.933 & 0.676 & 0.950 & \textbf{1.000} & 0.900 & \textbf{1.000} & 0.956 & 0.972 & 0.983 & 0.972 \\
PPOC & \textbf{0.904} & 0.876 & 0.628 & 0.828 & \textbf{0.921} & 0.852 & 0.907 & 0.900 & 0.900 & 0.897 & 0.790 \\
SHGCh2 & 0.853 & \textbf{0.898} & 0.623 & \textbf{0.898} & \textbf{0.955} & 0.853 & 0.960 & 0.770 & 0.953 & 0.950 & 0.897 \\
SS & \textbf{1.000} & \textbf{1.000} & 0.623 & \textbf{1.000} & \textbf{1.000} & 0.989 & 0.961 & \textbf{1.000} & \textbf{1.000} & \textbf{1.000} & 0.522 \\
Sony1 & \textbf{0.900} & 0.869 & 0.622 & 0.855 & 0.894 & 0.463 & 0.925 & 0.892 & 0.919 & 0.910 & 0.729 \\
Sony2 & \textbf{0.924} & \textbf{0.924} & 0.597 & 0.919 & \textbf{0.936} & 0.912 & 0.919 & 0.948 & 0.905 & 0.929 & 0.885 \\
Strawberry & \textbf{0.983} & 0.968 & 0.594 & 0.932 & 0.973 & 0.976 & 0.970 & 0.978 & 0.960 & 0.981 & 0.941 \\
ToeS1 & \textbf{0.939} & \textbf{0.939} & 0.587 & 0.877 & \textbf{0.952} & 0.947 & 0.877 & \textbf{0.974} & 0.838 & \textbf{0.969} & 0.737 \\
ToeS2 & \textbf{0.953} & \textbf{0.953} & 0.569 & 0.792 & 0.915 & \textbf{0.969} & 0.862 & 0.915 & 0.854 & 0.938 & 0.862 \\
TLECG & \textbf{0.998} & 0.985 & 0.500 & 0.826 & 0.998 & 0.997 & 0.987 & 0.998 & 0.994 & \textbf{0.999} & 0.848 \\
Wafer & \textbf{1.000} & 0.995 & \textbf{1.000} & 0.998 & 0.999 & 0.999 & \textbf{1.000} & \textbf{1.000} & \textbf{1.000} & \textbf{1.000} & 0.996 \\
Wine & \textbf{0.994} & 0.870 & 0.722 & 0.444 & 0.852 & 0.759 & 0.667 & 0.778 & 0.778 & 0.796 & 0.593 \\
WTC & \textbf{0.994} & 0.740 & 0.779 & 0.831 & 0.766 & 0.779 & 0.779 & 0.818 & 0.857 & 0.818 & 0.610 \\
\bottomrule
\end{tabular}
\label{tab: table2}
\end{table*}

\subsection{Simulation Study}
As defined in the paper, we represent a pattern using a sine curve with the formula:

$$
P(t) = A \cdot \sin(2 \pi f t + \phi)
$$

where \(A\) is the amplitude, \(f\) is the frequency, \(\phi\) is the phase shift, and \(t\) is the time index. The combination of patterns is:

$$
C(t) = \sum_{j=1}^k w_j \cdot P_{i_j}(t - \delta) + \mathcal{N}(0, \sigma).
$$

\begin{figure}[!ht]
\centering
\includegraphics[width=\textwidth]{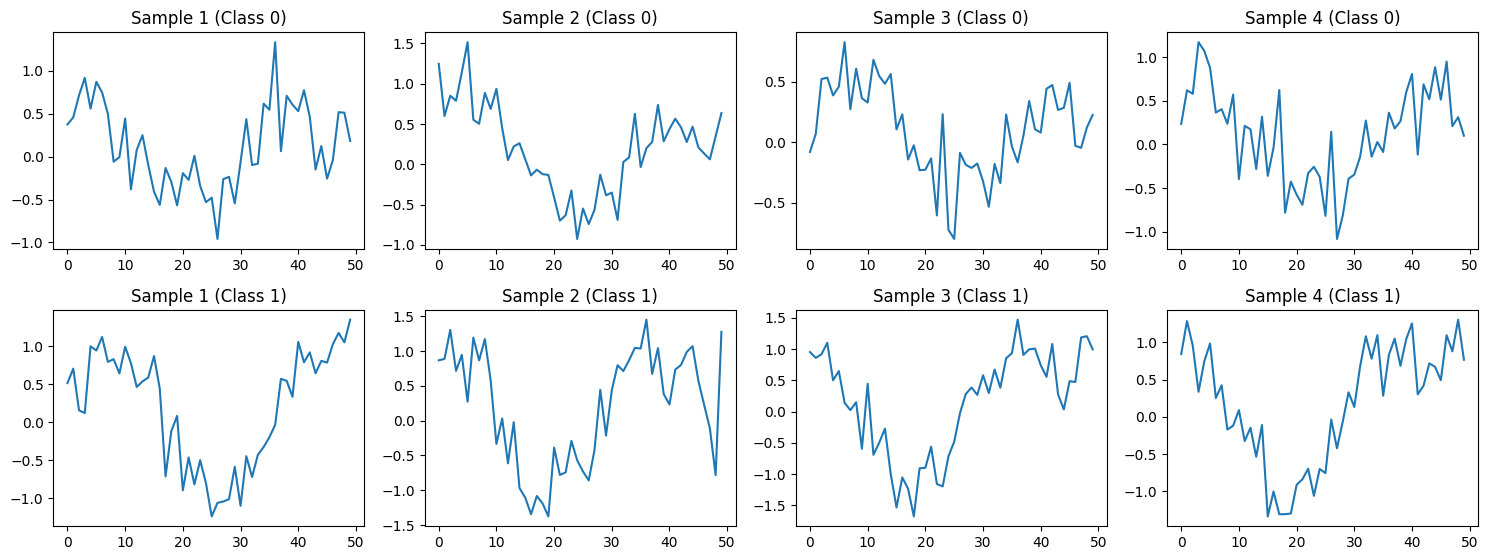}
\caption{A sample simulation plot.}
\label{fig: simu_sample}
\end{figure}

To simulate different levels of noise, multimodality, and class similarity, we use this model. A sample plot illustrating this is included in Figure \ref{fig: simu_sample}.

For noise levels ranging from 0 to 5, we add 0.1 to the baseline noise level, starting from a baseline \(\sigma\) of 0.3. For Class 0, we set the frequency range \(f \in (1.0, 1.5)\), amplitude range \(A \in (0.4, 1.2)\), and phase range \(\phi \in (0.0, 0.8)\). For Class 1, we set the frequency range \(f \in (1.1, 1.6)\), amplitude range \(A \in (0.5, 1.3)\), and phase range \(\phi \in (0.2, 1.0)\).

To simulate different class similarities, we define the most similar cases for level 5 with parameter settings the same as the baseline for noise. To decrease the level of similarity between the two classes, we subtract 0.2 from each interval range for Class 0 and add 0.2 to each interval range for Class 1.

Lastly, multimodality refers to cases where we have different patterns within each class, making classification challenging. From a level of 5 to 0, we simulate less multimodal distributions within each class by shrinking the parameter ranges by 0.1, starting from the most multimodal distributions defined the same as the baseline for simulating noise.

\clearpage
\newpage

\bibliographystyle{unsrt}  


\end{document}